\documentclass{svproc}
\usepackage{url}
\usepackage{times}
\usepackage{helvet}
\usepackage{courier}
\usepackage{fixltx2e}
\usepackage{graphicx}
\usepackage{etoolbox}
\usepackage{tabularx} 
\usepackage{amsmath}
\usepackage{amssymb}
\usepackage{inputenc}
\usepackage{verbatim}
\usepackage{mathtools}
\usepackage{float}
\usepackage{color}
\usepackage{balance}
\usepackage[linesnumbered]{algorithm2e}
\usepackage{comment}
\usepackage{dblfloatfix}

\begin{document}
\mainmatter              
\title{Moving Towards Open Set Incremental Learning: Readily Discovering New Authors}
\titlerunning{Moving Towards Incremental Learning}  
%
\author{Justin Leo\inst{1} \and Jugal Kalita\inst{2}}
\authorrunning{Leo and Kalita} 
%
%
\institute{University of Colorado, Colorado Springs CO 80918, USA\\
\email{jleo@uccs.edu},
\and
University of Colorado,
Colorado Springs CO 80918, USA\\
\email{jkalita@uccs.edu}}

\maketitle              

\begin{abstract}
The classification of textual data often yields important information. Most  classifiers work in a closed world setting where the classifier is trained on a known corpus, and then it is tested on unseen examples that belong to one of the classes seen during training. Despite the usefulness of this design, often there is a need to classify unseen examples that {\em do not belong} to any of the classes on which the classifier was trained. This paper describes the open set scenario where unseen examples from previously unseen classes are handled while testing. This further examines a process of enhanced open set classification with a deep neural network that discovers new classes by clustering the examples identified as belonging to unknown classes, followed by a process of retraining the classifier with newly recognized classes. Through this process the model moves to an incremental learning model where it continuously finds and learns from novel classes of data that have been identified automatically. This paper also develops a new metric that measures multiple attributes of clustering open set data. Multiple experiments across two author attribution data sets demonstrate the creation an incremental model that produces excellent results.

\end{abstract}
\section{Introduction}
Formal as well as informal textual data are over-abundant in this Internet-connected era of democratized publishing and writing. These textual information sources are in multiple forms such as news articles, electronic books and social media posts.  The use of text classification allows us to determine important information about the texts that can often be used to connect to the respective authors,  naturally  leading to the concept of Authorship Attribution. Authorship Attribution is seen as the process of accurately finding the author of a piece of text based on its stylistic characteristics \cite{rocha2016authorship}. Authorship Attribution is useful in scenarios such as identification of  the author of malicious texts or the analysis of historical works with unknown authors. 

Typically, text classification has a few well-established stages. The words in the text corpus are transformed using an embedding algorithm, and  a classifier is trained with  documents labeled with associated classes. In  Authorship Attribution, the text samples tend to be  books such as novels, transcribed speeches, or Internet-mediated social media posts, where each  sample is labeled with the corresponding author. The trained text classifier is given testing data that is usually unseen text samples from the same set of trained authors. This process describes a closed set approach because the tested samples are associated with the same trained classes. A problem with this process of classification arises if the testing data includes samples from unfamiliar authors. In these  cases, the classifier typically and erroneously associates the piece of text with a wrong author---an author on which it was trained. To remedy this problem, a new approach called open set classification has been proposed.  Open set classification enables the classifier  to discriminate among the  known classes, but additionally and  importantly,  to identify if some test example is not associated with any of  the  classes on which it was trained \cite{scheirer2012toward}.

There has been some recent work on open set classification using convolution neural networks (CNN) and recurrent neural networks (RNN). Prior work on open set classification has often been in areas such as computer vision \cite{bendale2015towards}, speech processing \cite{dahl2011context}, and natural language processing \cite{higashinaka2014towards}. This paper utilizes open set recognition to identify the presence of test examples from novel classes, and incorporate these new classes to those already known to create an incremental class-learning model.

The rest of the paper is organized as follows. After describing the related work in the the next section, the approach is presented to identify new classes and instantiate them. Then, the following section discusses evaluation metrics for assessing incremental learning, followed by experimental results using authorship attribution datasets and analysis. The final section reiterates the research accomplishments and thoughts on future work.   
\section{Related Work}
The discussed related work is in terms of four topics: deep networks for open set classification, metrics for open set classification, open set text classification, and recent proposals to use loss functions for open set classification in the context of computer vision. 

\subsection{Open Set Deep Networks}
Using deep neural networks for  open set classification often requires a change in the network model. Modern neural networks have multiple  layers connected in various ways, depending on the classifier architecture being used. Most  models eventually include a softmax layer that classifies the data to the known classes, with an associated confidence level or probability for each class.   A test example is considered to belong to the class which has the highest probability among all the classes. To adapt this model to the open set scenario,  the softmax layer was replaced by a unique  layer named the OpenMax   layer \cite{bendale2016towards}. This layer estimates the probability of an input being from one of the known classes  as well as an ``unknown" class, which lumps together all classes unseen during training. 
Thus, the network is able to recognize examples belonging to unknown classes, enhancing the ability of the closed set classifier it starts with. 

\subsection{Metric for Evaluating Open Set Classification}
The process of open set class recognition leads to new challenges during the evaluation process. There are multiple sources of error that could be present including: misclassification of known or unknown classes and determination of novel classes. Bendale and Boult (2015) proposed a metric to evaluate 
how individual examples are classified. Although the metric was originally proposed for use in computer vision, it could be applicable in author attribution as well. 

\subsection{Deep Open Set Text Classification}
Prakhya, Venkataram, and Kalita (2017)
modify the single OpenMax layer proposed by \cite{bendale2016towards} to replace the softmax layer  
in a multi-layer convolution neural networks with an ensemble of several  outlier detectors to obtain high accuracy scores for open set textual classification. The ensemble of classifiers uses a voting model between three different approaches: Mahalanobis Weibull, Local Outlier Factor \cite{kriegel2009loop}, and Isolation Forest \cite{liu2008isolation}. The average voting method produced results that are more accurate in detecting outliers, making detection of unknown classes better. 

\subsection{Loss Functions for Open Set Classification}
A problem that often occurs in open set classification is the classifier labeling known class data as unknown. This problem typically occurs if there are some similar features in the examples of the pre-trained classes and unknown classes encountered during testing. In the context of computer vision,   Dhamija, G{\"u}nther, and Boult (2018) introduce what is called the  Entropic Open-Set loss function  that increases the entropy of the softmax scores for background training samples and improves the handling of background and unknown inputs. They introduce  another loss  function called  the Objectosphere loss, which further increases softmax entropy and performance by reducing the vector magnitudes of examples of unknown classes in comparison with those from the known classes, lowering the erroneous classification of known class data as unknown. Since this approach squishes the magnitudes of all examples that belong to all unknown classes, it makes later separation of individual unknown classes difficult.
\section{Approach}

\begin{figure}[ht]
\centering
\includegraphics[width=10cm]{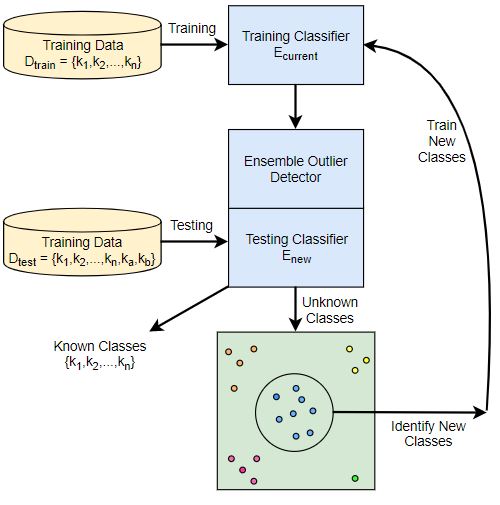}
\caption{Protocol for Open Set Classification and Incremental Class Learning}
\label{figure:protocol}
\end{figure}

\begin{figure}[ht]
\centering
\includegraphics[width=\linewidth]{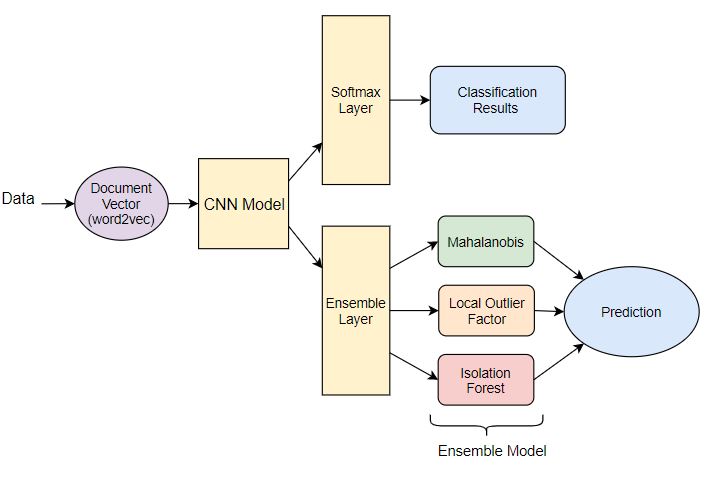}
\caption{Ensemble Model and Testing Classifier Diagram. This diagram more clearly describes the `Ensemble Outlier Detector' component from Figure \ref{figure:protocol}.}
\label{figure:protocol2}
\end{figure}

This paper explores open set classification and the process of moving towards incremental learning of new classes. The objective is to create a classifier framework that can incrementally learn and expand its knowledge base as additional data is presented as shown in Figure \ref{figure:protocol} and Figure \ref{figure:protocol2}. The approach is also outlined in Algorithm \ref{algorithm:protocol}.

In prior work on open set classification,  authors have focused on recognizing test samples as belonging to classes unknown during training. Based on prior research and knowledge, this paper is the first to instantiate new classes iteratively, extending prior work to real incremental class learning. Initially there is a summarization of the approach to provide an easily comprehensible sketch before moving on to details. The classifier framework is initially initialized by training it with examples from a small number of selected classes. The trained classifier is then exposed to a mix of examples from the already-known classes as well unknown classes, during testing. At a certain point, the testing of the current-classifier is paused and then all examples recognized as belonging to unknown classes are clustered. Clustering allows for the grouping of similar data and visually represents the differences between unique clusters. The hypothesis is that, if the clustering is good, one or more of the clusters of unknown examples can be thought of as new classes the current-classifier has not seen and these clusters are instantiated as new classes, by making up new unique labels for them. At this point, the current-classifier is updated by retraining it with all examples of the old known classes as well the newly instantiated classes. This process of training, accumulating of outliers, clustering, and instantiating selected new classes out of the clusters is repeated repeated a number of times as long as the error of the entire learning process remains acceptable. 
 
\begin{algorithm*}
\KwIn {Training Set ${\bf D} = \big<{\bf x}^{(i)}, y^{(i)} \big>, i=1\cdots N$, samples from all known classes}

\KwOut {An incrementally trained classifier $E$ on examples from a number of classes in $\bf D$   } 

$\mathcal{C}_{all} \leftarrow \big\{C_1, \cdots C_n \big\}$, set of all known classes

$\mathcal{C}_{current}^{train} \leftarrow$ (randomly) pick $k_{seed}$ classes from $\mathcal{C}_{all}$

${\bf D}_{current}^{train} \leftarrow {} \big\{ \big<{\bf x}^{(i)}, y^{(i)} \big>  \mid y{(i)} \in \mathcal{C}_{current}^{train}  \big\}$, samples from classes in $\mathcal{C}_{current}^{train}$

\Repeat{too low accuracy {\bf or} $n$ times}
{

$\mathcal{C}^{unknown}_{current} \leftarrow$ (randomly) pick $k_{unknown}$ classes from $\mathcal{C}_{all} - \mathcal{C}_{current}^{train}$

${\bf D}^{test}_{current} \rightarrow {\bf D}^{train}_{current} \bigcup 
        \big\{ \big<{\bf x}^{(i)}, y^{(i)} \big>  \mid y^{(i)} \in \mathcal{C}_{current}^{unknown}  \big\}$

$E_{current} \leftarrow $ (CNN) classifier trained on ${\bf D}_{currrent}^{train}$

${\bf O} \leftarrow$ outlier samples detected by ensemble outlier detector   when tested on ${\bf D}^{test}_{current}$

$\mathcal{L} \leftarrow $ set of clusters produced from $\bf O$ using a selected clustering algorithm

$\mathcal{L}_{dominant} \rightarrow $ pick $k_{new}$ dominant clusters from $\mathcal{L}$, call these clusters new classes by making up new labels for them

$\mathcal{C}_{current}^{train} \leftarrow\mathcal{C}_{current}^{train} \bigcup \mathcal{L}_{dominant} $, increase the number of ``known" classes

${\bf D}_{current}^{train} \leftarrow {\bf D}_{current}^{train} \bigcup 
    \big\{    
        \big<x,y \big> \in L_j \mid L_j \in \mathcal{L}_{dominant}
    \big\}$
    
}

$E \rightarrow E_{current}$

\Return {$E$}

\caption{Algorithm for Incremental Class-Learning}
\label{algorithm:protocol}
\end{algorithm*}

 In particular, the classifier is a multi-layer CNN structure for training purposes. During testing,  the  softmax layer at the very end replaced by an outlier ensemble, following the work of \cite{prakhya2017open}.  The outlier  detector ensemble consists of a  Mahalanobis model, Local Outlier Factor model, and an Isolation Forest model, like \cite{prakhya2017open}. The classifier model, as used in training is shown in Figure \ref{figure:protocol2}. Initially the model is created by training a  classifier $E_{current}$ with a given $k_{seed}$ number of classes found in the entire training data set $\bf D$. Then a derived dataset is created $D_{current}^{test}$  for testing the model by mixing examples of $k_{unknown}$  unknown classes with the previously trained $k_{seed}$ classes. 
 The process always adds $k_{new}$ classes to the number of known classes. Thus, at the end of the $i$th iteration of class-learning, the classifier knows $k_{seed} + (i-1) k_{new}$ classes. The model instantiates ``new" classes by choosing dominant clusters, and then retrain the model with these new classes. The classes are then removed from the set of all classes and new ones are selected for the incremental addition.

 This paper experiments with multiple clustering techniques including K-Means \cite{hartigan1979algorithm}, Birch \cite{zhang1996birch}, DBScan \cite{ester1996density}, and Spectral \cite{stella2003multiclass}, to determine the  most suitable one for author attribution. There is also experimentation with various values of the parameters: $k_{seed}$, $k_{unknown}$ and $\delta$.

\section{Evaluation Methods}
Since the method uses clustering as well as classification in the designed protocol for incremental classification, there needs to be evaluation of both. First, it is required to have an outline of how clusters obtained from examples classified as unknown are evaluated, and then it is required to have a description of how the incremental classifier is evaluated. 

\subsection{Evaluation of Clustering}
There are a variety of clustering algorithms, and the model needs one that works efficiently in the domain of author attribution. The test samples that are deemed to be outliers are clustered, with the hypothesis that some of these clusters correspond to actual classes in the original dataset. 
The evaluation process uses the Davies-Bouldin Index as shown in Equation (\ref{equation:DaviesBouldin})  to evaluate clustering \cite{davies1979cluster}.
\begin{equation} 
    DB = \frac{1}{n} \sum_{i=1}^{n}{max}_{j\neq i}
    \bigg(\frac{\sigma\textsubscript{i} + \sigma\textsubscript{j}}{d(c\textsubscript{i}, c\textsubscript{j})}
    \bigg)
\label{equation:DaviesBouldin}
\end{equation}
In this formula, $n$ is the number of clusters produced, $\sigma_i$ is the average distance between the points in cluster $i$ and its centroid, $d(c_i, c_j)$ is the Euclidean distance between the centroids of clusters indexed $i$ and $j$. 
Typically lower Davies-Bouldin Index scores indicate better clustering. Another  clustering evaluation metric used is the V-Measure as shown in Equation (\ref{equation:VMeasure}), which has been widely used in clustering in natural language processing tasks when ground truth is known, i.e., the samples and their corresponding classes are  known. This metric  computes the harmonic mean between homogeneity and completeness \cite{rosenberg2007v}. Homogeneity measures how close the clustering is such that each cluster contains samples from one class only. Completeness measures how close the clustering is such that samples of a given class are assigned to the same cluster. 
Typically scores close to 1 indicate better clustering. Here $\beta$ is a parameter used to weigh between the two components---a higher value of $\beta$ weighs completeness more heavily over homogeneity, and vice versa. 

\begin{equation} 
    V = \frac{(1 + \beta) *homogeneity * completeness}
        {\beta * homogeneity + completeness}
\label{equation:VMeasure}
\end{equation}

\subsection{Evaluation of Open Set Misclassification Error}
Assuming there are $n$ known classes, multi-class classification using a classifier $E_n()$, trained on $n$ classes, can be evaluated using the misclassification error: 
\begin{equation}
    \epsilon_n = \frac{1}{N}
      \sum_{i=1}^N  
        \big[ 
        E_n ({\bf x}^{(i)}) \neq y^{(i)}
        \big]
\end{equation}
where $N$ is the total number of samples in the dataset. When the same classifier $E_n()$ is tested in the context of open set classification, there is a need to keep track of errors due that occur between known and unknown classes. When the classifier is tested on $N$ samples from $n$ known classes and $N^\prime$ samples from $u$ unknown classes, the test is a total of $N+N^\prime$ samples over $n+u$ classes. The open set classification error $\epsilon_{OS}$ for classifier $E_n$ is given as \cite{bendale2015towards}:
\begin{equation}
    \epsilon_{OS} = \epsilon_n + 
     \frac{1}{N^\prime}
     \sum_{j=N+1}^{N^\prime} 
        \big[ 
        E_n ({\bf x}^{(i)}) \neq unknown 
        \big]
\end{equation}

\subsection{Evaluation of Incremental Class Learning Accuracy}
For this research, the approach uses clustering in order to obtain new classes after open set recognition is performed.
This way the new data identified for the novel classes can be used to incrementally train the model. For the evaluation of these clusters this paper presents a new metric \textit{ICA} (Incremental Class Accuracy) which takes into account the specific data from an identified cluster and averages calculations of homogeneity, completeness, and unknown identification accuracy of the cluster. This paper defines homogeneity as the ratio of the number of data samples of the predominant class \textit{c} in the cluster \textit{k} (\textit{n\textsubscript{c$\mid$k}}) and the total number of values in the cluster (\textit{N\textsubscript{k}}). This paper defines defines completeness as the ratio of the number of data samples of the predominant class \textit{c} in the cluster \textit{k} (\textit{n\textsubscript{c$\mid$k}}) and the total number of tested samples of the same class \textit{N\textsubscript{c}}. This paper defines define unknown identification accuracy as ratio of the number of unknown \textit{u} data samples in the cluster \textit{k} (\textit{n\textsubscript{u$\mid$k}}) and the total number on values in the cluster \textit{N\textsubscript{k}}. The equation used for ICA assumes only one cluster is being evaluated, but the equation can be adapted for multiple clustering by finding multiple ICA scores for each cluster and averaging. 

\begin{equation}
    Homogeneity = \frac{max(n\textsubscript{c$\mid$k})}{N\textsubscript{k}}
\end{equation}
\begin{equation}
    Completeness = \frac{max(n\textsubscript{c$\mid$k})}{N\textsubscript{c}}
\end{equation}
\begin{equation}
    \mbox{\textit{Unknown Identification Accuracy}} = \frac{(n\textsubscript{u$\mid$k})}{N\textsubscript{k}}
\end{equation}
\begin{equation} 
    ICA = \Bigg(\frac{max(n\textsubscript{c$\mid$k})}{N\textsubscript{k}}+
    \frac{max(n\textsubscript{c$\mid$k})}{N\textsubscript{c}}+
    \frac{(n\textsubscript{u$\mid$k})}{N\textsubscript{k}}\Bigg)*
    \frac{1}{3}
\end{equation}


Other metrics that will  be used to determine the performance of the model will be accuracy and F1-score, these figures inherently show the accuracy of the classifier as well as novel data detection.

\section{Experiments and Results}
This section discusses the data sets used, the experiments performed, and the results with analysis. 

\subsection{Datasets}
Since the objective is for open set author attribution, the testing uses two datasets each of which contains 50 authors. 
\begin{itemize}
    \item \textbf{Victorian Era Literature Data Set} \cite{gungor2018benchmarking}: This dataset is a collection of writing excerpts from 50 Victorian authors chosen from the GDELT database. The text has been pre-processed to remove specific words that identify the individual piece of text or author (names, author made words, etc.). Each author has hundreds of unique text pieces with 1000 words each. 
    \item \textbf{CCAT-50} \cite{houvardas2006n}: This data set is a collection of 50 authors each with 50 unique text pieces divided for both training and testing. These texts are collections of corporate and industrial company news stories. This data is a subset of Reuters Corpus Volume 1.
\end{itemize}

\subsection{Preliminary Clustering Results}
\begin{figure*}[ht]
\centering
\includegraphics[width=\textwidth]{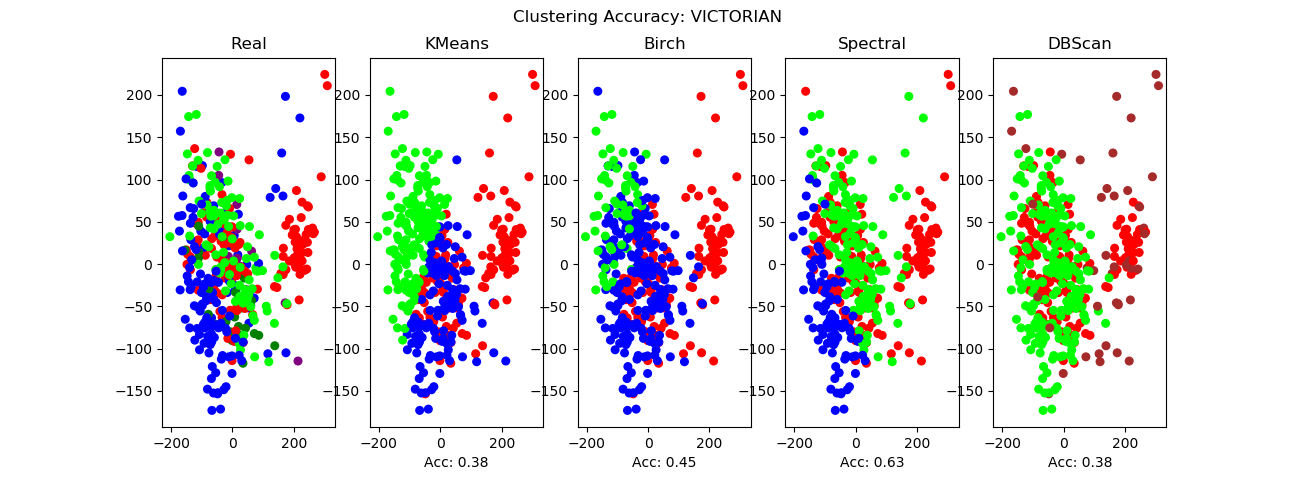}
\caption{Clustering Plots for Victorian Literature Data with Accuracy Score, 5 Trained Classes and 8 Tested Classes}
\label{figure:clusteringPlotsVictorian}
\includegraphics[width=\textwidth]{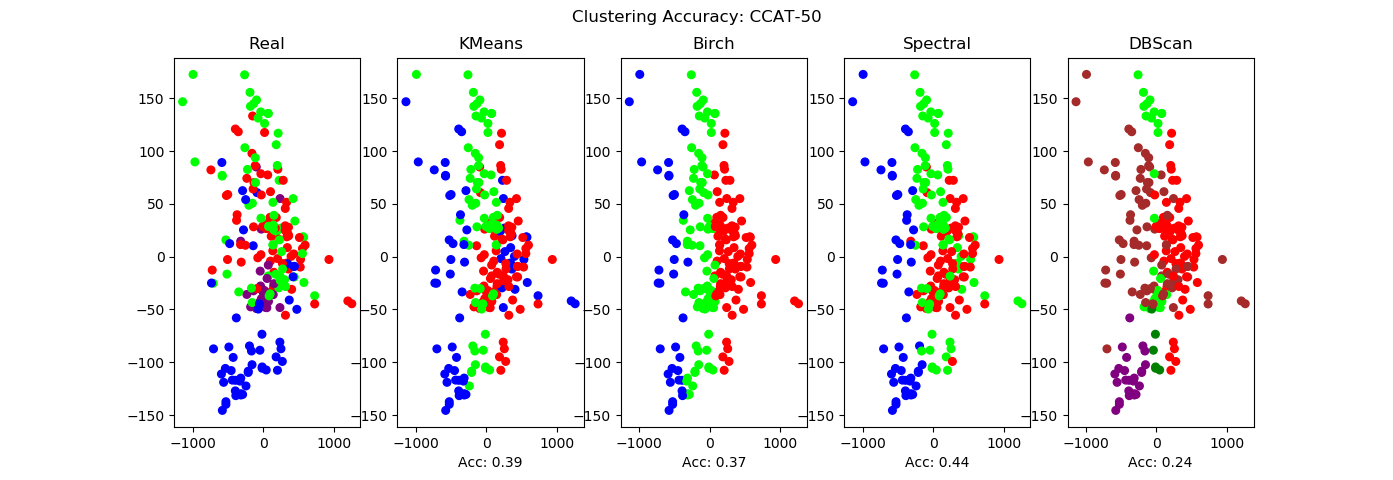}
\caption{Clustering Plots for CCAT-50 Data with Accuracy Score, 5 Trained Classes and 8 Tested Classes}
\label{figure:clusteringPlotsCCAT}
\end{figure*}

After experimental comparison of the different clustering techniques, the final decision was to use Spectral Clustering \cite{stella2003multiclass} as this typically produces the highest accuracy results as seen in Figure \ref{figure:clusteringPlotsVictorian} and Figure \ref{figure:clusteringPlotsCCAT}, the clustering evaluation scores are also used for comparison. The pre-trained model \textit{word2vec} \cite{mikolov2013distributed} to obtain the word embeddings to pass into the multi-layer CNN structure. 

\subsection{Incremental Classification Results}
For the first experiment the objective was to see if the proposed method would improve the calculated classification accuracy and to also decide which clustering algorithm would work best. 
Both data sets are run individually with five known training classes and then with ten known training classes, then the model is introduced to three unknown classes during the testing phase for each of the tests.
The results include the comparison with accuracy and F1-Score as found on Table 1; a significant increase of these values is observed after the classifier is retrained with the identified novel classes. The clustering evaluation metrics are found on Table 2.
V-Measure scores prove to be more useful because the Davies-Bouldin scores do not always indicate the highest accuracy of clustering, this is because the best formed clusters does not necessarily mean higher accuracy.
Even though the chosen data sets have not been used for open set classification in prior research this paper compares the calculated open set classification scores with the state of the art closed set classification scores.
Based on prior research, the best classification F1-Score from prior work for the Victorian Literature data set using only few classes is 0.808 \cite{gungor2018benchmarking} and the designed model produces a slightly better score.
Also based on prior research, the best classification accuracy score for the CCAT-50 data set for using only few classes is 86.5\% and the designed model obtains similar results.
The clustering models seem to have the most error for both data sets (especially the CCAT-50 data), thus presumably better clustering models or would produce greater results. 

\begin{table}[ht]
  \centering
  \renewcommand{\arraystretch}{1.2}
  \begin{tabular}{|p{2.5cm}||c|c|c|c|}
    \hline
    Dataset & \multicolumn{2}{c|}{Pre-Trained} & \multicolumn{2}{c|}{Post-OpenSet}\\
    \cline{2-5}
    & Acc & F1 & Acc & F1 \\
    \hline
    Victorian 5class & 56.29\% & 0.592 & 85.43\% & 0.855 \\
    CCAT-50 5class & 54.75\% & 0.565 & 83.00\% & 0.825 \\ 
    Victorian 10class & 61.29\% & 0.644 & 71.38\% & 0.706  \\
    CCAT-50 10class & 62.50\% & 0.727 & 86.77\% & 0.866  \\ 
    \hline
  \end{tabular}
  \caption{Pre-Trained Class Scores and Post-Open Set Classification Scores, Either 5 or 10 initial trained classes and 3 unknown added during testing}
\end{table}

\begin{table}[ht]
  \centering
  \renewcommand{\arraystretch}{1.2}
  \begin{tabular}{|p{2.5cm}||c|c|c|c|c|c|c|c|}
    \hline
     & \multicolumn{4}{c|}{Davies-Bouldin} & \multicolumn{4}{c|}{V-Measure}\\
    \cline{1-9}
    Data Set& Vic-5 & CCAT-5 & Vic-10 & CCAT-10 & Vic-5 & CCAT-5 & Vic-10 & CCAT-10 \\
    \hline
    K-Means & 2.739 & 2.045 & 1.989 & 0.876 & 0.078 & 0.039 & 0.147 & 0.082\\
    Birch & 2.670 & 2.237 & 4.193 & 3.654 & 0.165 & 0.075 & 0.147 & 0.083\\
    Spectral & 4.457 & 2.550 & 4.841 & 0.807 & 0.319 & 0.242 & 0.328 & 0.258\\
    DBScan & 4.031 & 2.432 & 4.783 & 4.349 & 0.065 & 0.101 & 0.158 & 0.149\\
    \hline
  \end{tabular}
  \caption{Davies Bouldin Index and V-Measure Score for Clustering methods evaluated, Either 5 or 10 trained classes and 3 unknown added during testing.}
\end{table}

\par
\begin{figure*}[ht]
\centering
\includegraphics[width=\textwidth]{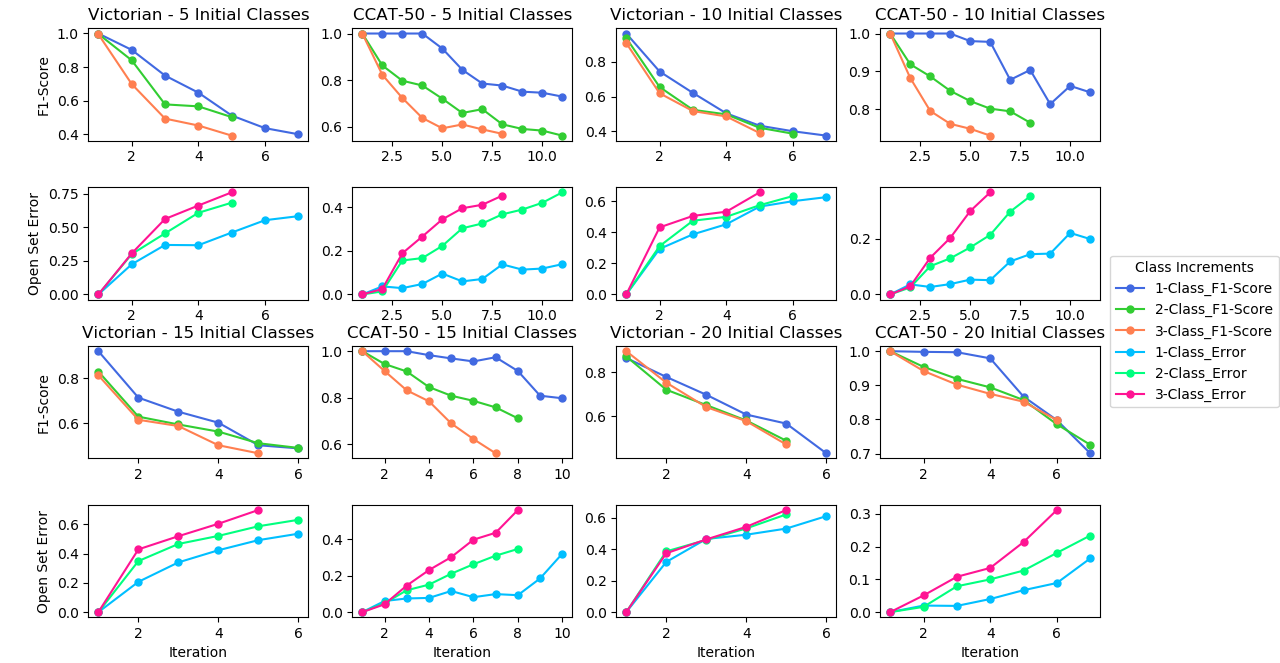}
\caption{Incremental Learning Plots. Initially trained with 5, 10, 15, and 20 initial classes then tested by incrementally adding 1, 2, and 3 Classes. These plots show the final F1-Scores and Open Set Error from Equation 4.}
\label{figure:IncrementalLearningResults}
\end{figure*}

For the second experiment the model is initially trained with a fixed amount of classes $k_{seed}$ and then the method incrementally adds a $k_{unknown}$ amount of classes for testing. This process is repeated to demonstrate the model incrementally learns as the learning and open set classification cycle is repeated. This test is run by adding classes for multiple iterations and record the change in the F1-Score for the overall classification and generation of new classes; the objective is to run each test until the results drop significantly or until the model reaches a max value of classes. Figure \ref{figure:IncrementalLearningResults} displays the results of the incremental cycle, and it is observed that the model achieves better results when fewer classes are added at a time. The experiment runs tests for adding 1, 2, and 3 classes at a time. The open set error shown in Equation 4 is also calculated for each test; this metric shows error of unknown data identification but not novel class generation. The problem noticed with the experiment is that error will propagate through the process so as error accumulates the results deter. Another observation, based on the results from both data sets, is that adding one class incrementally each iteration has better results because this limits the clustering error. It is also clear that the Victorian Literature performs worse than the CCAT-50 data and the initial reasoning for this is because of the text samples; the Victorian text includes words with slurs and accent mark symbols and \textit{word2vec} is not pre-trained with these new features. The CCAT-50 data tends to have very distinct authors and the pieces of text tend to also tend to be more unique. Overall based on the results, it is concluded that most of this error can be attributed to the clustering process.

\par
As stated in the previous experiments, the clustering process tends to have the most variance, this is evident from the low clustering accuracy due to the lack of fully distinct clusters. Thus, there needs to be a way to evaluate the clustering. Using the Incremental Class Accuracy (ICA) metric shown from Equation 5, there will be ability to evaluate the clustering in regards to homogeneity, completeness, and unknown identification accuracy. From the previous experiment it is also noticed that adding one class at a time incrementally tends to produce the best results, so the ICA score is calculated when one class is added and instantiated. The results for both data sets is shown in Table 3. From these results it is observed that having a fewer amount of initial trained $k_{seed}$ classes produces better results and this is expected as the $k_{unknown}$ classes are more easily identified.
\begin{table}[H]
  \centering
  \renewcommand{\arraystretch}{1.2}
  \begin{tabular}{|p{2.5cm}||c|c|}
    \hline
     Initial Training & Victorian & CCAT-50 \\
    \hline
    5 Classes & 0.687 & 0.875\\
    10 Classes & 0.593 & 0.754 \\
    15 Classes & 0.529 & 0.764 \\
    20 Classes & 0.387 & 0.681 \\
    \hline
  \end{tabular}
  \caption{ICA Scores for 1 added class/cluster evaluation. Scores based on Equation 5.}
\end{table}

\section{Conclusion}
This research works with open set classification regarding NLP text analysis in the area of Authorship Attribution. The model created will be to determine the originating author for a piece of text based on textual characteristics. This research also move towards a novel incremental learning approach where unknown authors are identified and then the data is labeled so the classifier expands on its knowledge. Through this process there is expansion upon the state of the art implementation by creating a full cycle model by training on given data and then expanding the trained knowledge based on new data found for future testing. 
\par
Text based Authorship Attribution can be applied to research involving security and linguistic analysis. Some similar developing work using similar research methods involving image recognition \cite{rebuffi2017icarl}, this can be applied to facial recognition tasks and video surveillance applications. This model can also be further improved by developing a more precise way of distinguishing different pieces of text. Another method for future research is using backpropagation. Once novel classes are identified, the model should be then able to modify the already trained classifier with the $D_{current}^{train}$ data. Then the model can be tested with the $D_{current}^{test}$ to determine if the model can recognize previously unknown classes. Backpropagation of a neural network requires a fully inter connected set of layers that allow the processing of data through either side of the model \cite{hecht1992theory}. This process would save the step of fully retraining the classifier model. A similar approach to this can also be to add new "neurons" to a deep neural network to allow for an extension of a trained model \cite{draelos2017neurogenesis}. With these new future improvements the designed model can be further improved and potentially obtain better results.
\section{Acknowledgement}
The work reported in this paper is supported by the National Science Foundation under Grant No. 1659788. Any opinions, findings and conclusions or recommendations expressed in this work are those of the author(s) and do not necessarily reflect the views of the National Science Foundation.

%
%
\bibliographystyle{unsrt}

\end{document}